\documentclass{article}

\usepackage{PRIMEarxiv}

\usepackage[utf8]{inputenc} 
\usepackage[T1]{fontenc}    
\usepackage{hyperref}       
\usepackage{url}            
\usepackage{booktabs}       
\usepackage{amsfonts}       
\usepackage{nicefrac}       
\usepackage{microtype}      
\usepackage{lipsum}
\usepackage{fancyhdr}       
\usepackage{graphicx}       
\usepackage{multirow}
\usepackage{amsmath}
\usepackage{xcolor}
\graphicspath{{media/}}     
\usepackage{tabularx}
\usepackage{color}
\usepackage[vlined,boxed,commentsnumbered,linesnumbered,ruled]{algorithm2e}
\usepackage{adjustbox}
\usepackage{colortbl}
\usepackage{listings}
\definecolor{green}{RGB}{113,165,55}
\definecolor{blue}{RGB}{1,158,213}
\definecolor{red}{RGB}{220,10,10}
\definecolor{LightGreen}{HTML}{d4edda}  
\definecolor{LightBlue}{HTML}{d1ecf1}   
\usepackage[capitalize]{cleveref}
\Crefname{section}{Section}{Sections}
\Crefname{table}{Table}{Tables}
\Crefname{figure}{Figure}{Figures}
\Crefname{equation}{Equation}{Equations}
\Crefname{paragraph}{Paragraph}{Paragraphs}

\title{Learning Where to Embed: Noise-Aware Positional Embedding for Query Retrieval in Small-Object Detection}

\author{
\textbf{Yangchen Zeng$^{1,\dagger}$ \quad
Zhenyu Yu$^{2,\dagger}$ \quad
Dongming Jiang$^{3}$ \quad
Wenbo Zhang$^{4}$} \\
\textbf{Yifan Hong$^{5}$ \quad
Zhanhua Hu$^{6}$ \quad
Jiao Luo$^{7}$ \quad
Kangning Cui$^{8,9,*}$} \\[0.6em]
$^{1}$Southeast University \quad
$^{2}$Fudan University \quad
$^{3}$The University of Texas at Dallas \quad
$^{4}$Zhejiang Normal University \\
$^{5}$Data Space Research Institute, Hefei Comprehensive National Science Center \quad
$^{6}$Rice University \\
$^{7}$Huazhong Agricultural University \quad
$^{8}$City University of Hong Kong (Dongguan) \quad
$^{9}$Wake Forest University
}
\begin{document}
\maketitle
\begingroup
\renewcommand{\thefootnote}{\fnsymbol{footnote}}
\footnotetext[1]{Corresponding author: \texttt{cuij@wfu.edu}. \quad $^\dagger$ Equal contribution.}
\endgroup

\begin{abstract}
    Transformer-based detectors have advanced small-object detection, but they often remain inefficient and vulnerable to background-induced query noise, which motivates deep decoders to refine low-quality queries. We present HELP (\textbf{H}eatmap-guided \textbf{E}mbedding \textbf{L}earning \textbf{P}aradigm), a noise-aware positional-semantic fusion framework that studies \emph{where to embed} positional information by selectively preserving positional encodings in foreground-salient regions while suppressing background clutter. Within HELP, we introduce Heatmap-guided Positional Embedding (HPE) as the core embedding mechanism and visualize it with a heatbar for interpretable diagnosis and fine-tuning. HPE is integrated into both the encoder and decoder: it guides noise-suppressed feature encoding by injecting heatmap-aware positional encoding, and it enables high-quality query retrieval by filtering background-dominant embeddings via a gradient-based mask filter before decoding. To address feature sparsity in complex small targets, we integrate Linear-Snake Convolution to enrich retrieval-relevant representations. The gradient-based heatmap supervision is used during training only, incurring no additional gradient computation at inference. As a result, our design reduces decoder layers from eight to three and achieves a 59.4\% parameter reduction (66.3M vs.\ 163M) while maintaining consistent accuracy gains under a reduced compute budget across benchmarks. \textbf{Code Repository}: {\url{https://github.com/yidimopozhibai/Noise-Suppressed-Query-Retrieval}.}
    
\end{abstract}

\keywords{Query retrieval, Positional embedding, Small object detection}

\section{Introduction}
\label{sec:intro}

Small-object detection remains a challenging and fundamental task in computer vision, especially in aerial and remote sensing imagery where targets occupy few pixels and are easily obscured by cluttered backgrounds. These settings often involve extreme scale variation, dense object layouts, and a large portion of background, all of which degrade localization and recognition performance~\cite{wei2024review, yang2022scrdet++, li2025efficient, hou2020mdpcaps, cui2024palmprobnet, cui2024superpixel}. The growing demand for reliable detection in large-scale aerial scenes further highlights the need for methods that generalize across diverse imaging conditions and object densities~\cite{cheng2023towards, cui2025efficient, akyon2022sahi,jiang2026magma}.

Transformer-based detectors have attracted increasing attention for small-object detection because their global context modeling and set-based objective reduce reliance on dense proposals and heuristic post-processing~\cite{zhao2024detrs, zhang2022dino, wang2026dinov3, zhu2025orthomosaics}. However, their performance in cluttered aerial scenes is often limited by query qualities used for decoding. In particular, many query generation pipelines do not fully exploit readily available detection cues, including spatial position, class confidence, and bounding box regression signals, especially under limited training data. This leads to weak alignment between positional encodings and detection semantics, as positional information is not selectively emphasized in regions that are truly informative for small-object localization~\cite{meng2021conditional, li2022dn,jiang2026anatomy}.

A second limitation is that queries are easily polluted by background responses when the background portion is high. Low-quality background regions can still introduce nontrivial embeddings, yet most detectors lack an explicit mechanism to suppress such background induced query noise before decoding, which degrades both localization and classification performance~\cite{zhang2022dino, li2022dn, zhang2026center, chen2024implicit, chen2025blind}. This also creates an efficiency issue, since the decoder often requires multiple layers to repeatedly correct and refine suboptimal queries, increasing training and inference cost as attention complexity grows with token length and decoder depth~\cite{zhao2024detrs, jiangsada2025}. Small-object detection further intensifies these issues because informative features are sparse and fine-grained. Without dedicated feature engineering to strengthen sparse discriminative patterns, the representations supporting query retrieval remain fragile in cluttered scenes, which limit robustness and generalization~\cite{wei2024review, di2025toward, han2026seaf}.

To address these challenges, we propose \textbf{HELP} (Heatmap-guided Embedding Learning Paradigm), an efficient positional and semantic fusion framework that learns \emph{where to embed} positional information. Instead of uniformly injecting positional encodings across all spatial locations, HELP uses Heatmap-guided adaptive learning to selectively preserve positional encodings in foreground-salient regions and suppress them in background-dominant regions. In doing so, HELP couples positional cues with category and bounding box information, which improves the quality of representations used for query decoding and enhances robustness to background noise. Within HELP, we introduce \textbf{Heatmap-guided Positional Embedding (HPE)} mechanism, as well as a heatbar-based visualization of HPE that provides an interpretable view of embedding allocation and practical guidance for fine-tuning.

Building on HPE, we develop \textbf{MOHFE} (Multi-Scale ObjectBox-Heatmap Fusion Encoder) and \textbf{HQ-Retrieval} (High-Quality Query Retrieval). MOHFE performs noise-suppressed feature encoding by integrating class and bounding box semantics into multi-scale Heatmap-guided Positional Embeddings to strength the encoder-side alignment between positional cues and detection semantics. HQ-Retrieval generates discriminative queries and filters low-quality background queries, which reduces repeated queries and enables a much shallower decoder. This design substantially accelerates both training and inference. To address feature sparsity in complex small targets, we propose \textbf{LSConv} (Linear-Snake Convolution) to capture sparse discriminative patterns and improve query generation.

Our main contributions are summarized as follows:

\begin{itemize}
    \item We propose \textbf{HPE} with heatbar visualization for interpretable analysis and fine-tuning of selective positional embedding to enable noise-robust query decoding.
    \item We design \textbf{MOHFE} and \textbf{HQ-Retrieval} for noise-suppressed encoding and discriminative query retrieval via a gradient-based filter, which reduce decoder depth from eight to three.
    \item We integrate \textbf{LSConv} to strengthen sparse feature extraction, achieving a 59.4\% parameter reduction (66.3M vs.\ 163M) while accelerating training and inference.
    \item We validate our approach on five benchmarks and show consistent gains and good scalability.

\end{itemize}

\section{Related Work}

\subsection{CNN-based Small Object Detection}
CNN-based detectors have long been the dominant paradigm in object detection, benefiting from hierarchical feature extraction and multi-scale representations~\cite{he2016deep, pan2024accurate}. Representative one-stage methods~\cite{liu2016ssd, redmon2018yolov3, zyc2022behavior, sun2025bmdnet, wang2026cott, lin2017focal} prioritize efficiency, while two-stage approaches~\cite{ren2015faster, he2017mask} leverage region proposal mechanisms for improved localization. These designs have been extended to aeriel and remote sensing imagery, where small objects are common and multi-scale cues are crucial~\cite{hou2025fedc, han2024luffd}. For example, SuperYOLO~\cite{zhang2023superyolo} enhances small-object representations via super-resolution, and PCNet~\cite{cao2023pcnet} improves discrimination through comparative feature learning. Despite their strong performance, most CNN-based pipelines rely on dense candidate generation and subsequent post-processing with non maximum suppression (NMS)~\cite{hosang2017learning}. NMS introduces thresholding and greedy selection, which can be sensitive to scene statistics and may reduce cross-dataset generalization~\cite{cai2018cascade, cui2025detection, ren2015faster, hou2025fedc}. Moreover, dense candidate processing can incur substantial redundancy in high-resolution aerial imagery, where numerous small targets and large background regions significantly increase both training and inference cost.

\subsection{Transformer-based Small Object Detection}
Detectors with transformer backbones (DETRs) provide an alternative by adopting set prediction objectives and query-driven decoding, which replace NMS with global context modeling~\cite{carion2020end}. DETR-style frameworks and their variants~\cite{zhu2020deformable, meng2021conditional, zhang2022dino, zhao2024detrs} often combine convolutional backbones for feature extraction, Transformer encoders for context aggregation, and learnable queries for parallel decoding. Concurrently, efficiency-oriented Transformer designs such as DynamicViT~\cite{rao2021dynamicvit} reduce attention cost through dynamic token pruning and allocation. Several works improve small-object detection by using strategies such as focal loss with class activation map to mitigate class imbalance and improve localization, as well as sparse attention over target-relevant windows to accelerate detection~\cite{ma2024coarse, li2024sparseformer, tang2024optimized}. Nevertheless, the computational burden of vanilla self-attention on high-resolution inputs remains prohibitive. Prior adaptations include window-based attention~\cite{liu2021swin}, hierarchical or downsampling designs~\cite{wang2021pyramid,wu2021cvt}, and low-rank approximations~\cite{xiong2021nystromformer}. Complementary sparsification strategies have also been applied at the level of tokens, attention heads, and Transformer blocks~\cite{rao2021dynamicvit, shao2026flashsvd,he2024query}. While these approaches reduce computation, they may weaken fine-grained localization or introduce additional optimization complexity for small and densely distributed targets.

\begin{figure*}[t]
    \centering
    \includegraphics[width=\textwidth]{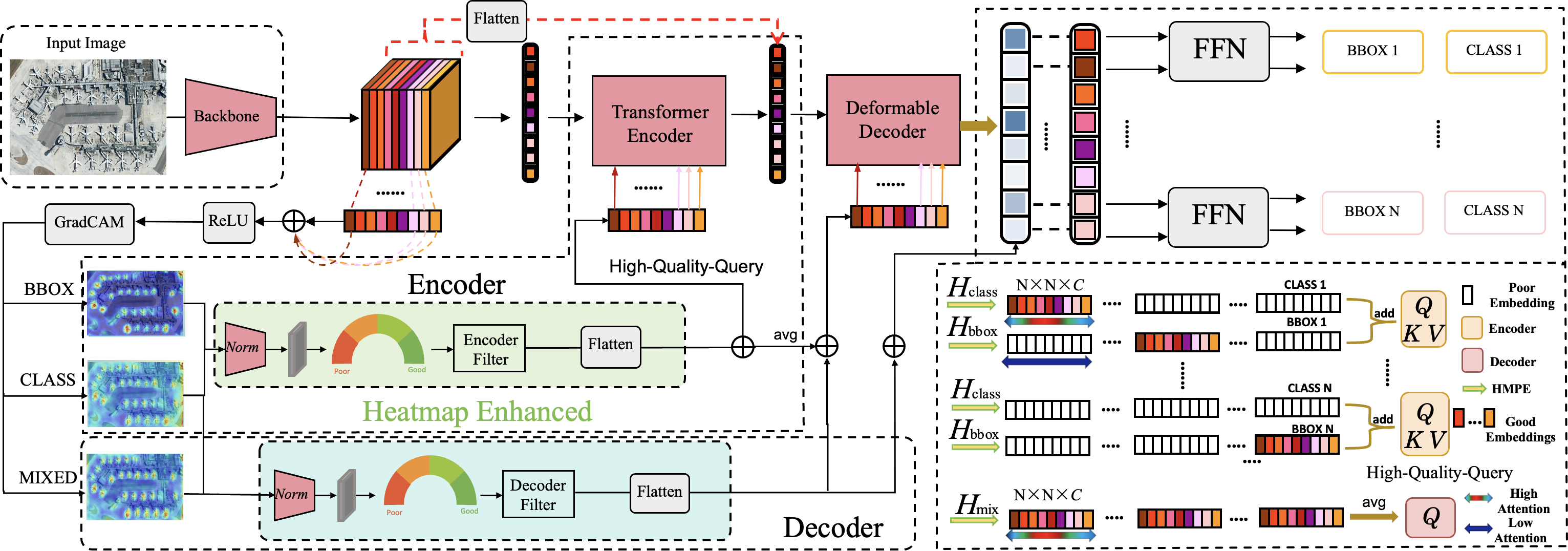}
    \caption{Overview of HELP.
    Backbone: LSConv is integrated into the backbone to enhance sparse features before transformer encoding.
    Encoder: MOHFE (green) fuses heatmap information with class and bounding box information to form heatmap-enhanced embeddings.
    Decoder: HQ-Retrieval (blue) selects informative embeddings for decoding while suppressing background-dominant ones.
    Bottom-right: HPE in action, where heatmap activation is used to select high-quality embeddings and filter out background-dominant ones.}
    \label{fig:ppl-crop}
\end{figure*}

\section{Heatmap-guided Embedding Learning}

This section presents the Heatmap-guided Embedding Learning Paradigm (HELP).
We first introduce Heatmap-guided Positional Embedding (HPE), which learns \emph{where} to preserve or suppress positional information based on heatmap-derived saliency.
We then describe how HPE is used in both the encoder and decoder for noise-suppressed feature encoding and high-quality query retrieval.
Finally, we present Linear-Snake Convolution (LSConv) as a complementary module that enriches sparse, retrieval-relevant representations for complex small-object scenarios.

\subsection{Heatmap-guided Positional Embedding}
\label{subsec: Heatmap-guided Positional Embedding （HPE ）and Visualization}

Within the proposed framework HELP, we introduce Heatmap-guided Positional Embedding (HPE) as the core embedding mechanism for noise-suppressed query retrieval.
As illustrated in Figure~\ref{fig:ppl-crop}, HPE learns \emph{where to embed} positional information by preserving positional cues in foreground-salient regions while suppressing background-dominant regions through the below steps:

\vspace{2mm}\noindent\emph{Step 1.} We extract intermediate feature tensors $\mathbf{A} \in \mathbb{R}^{K \times H \times W}$, where $A_k(i, j)$ denotes the activation of the $k$-th channel at spatial coordinate $(i, j)$.
For a detection class $c$, the gradient weighting coefficients $\alpha_{ijk}^{c}$ are derived by computing second- and third-order partial derivatives of the classification confidence $y^c$ with respect to $\mathbf{A}$:
\begin{equation}
    \alpha_{ijk}^{c} = \frac{\partial^2 y^c}{\partial A_k(i,j)^2} + \frac{\partial^3 y^c}{\partial A_k(i,j)^3}
\end{equation}
The channel-wise importance weight $\beta_{k}^{c}$ is computed via spatial aggregation with ReLU-based gradient filtering:
\begin{equation}
    \beta_{k}^{c} = \sum_{i=1}^{H} \sum_{j=1}^{W} \alpha_{ijk}^{c} \cdot \text{ReLU}\left( \frac{\partial y^c}{\partial A_k(i,j)} \right)
\end{equation}
Here, $\text{ReLU}$ suppresses negative gradients to retain only activations positively correlated with class $c$, while $\alpha_{ijk}^{c}$ amplifies regions with higher-order nonlinear contributions to discrimination.

\vspace{2mm}\noindent\emph{Step 2.} Using the channel-wise importance weights $\beta_k^c$, we compute the class-discriminative heatmap by:
\begin{equation}
    H_{\text{class}}(i,j) = \text{ReLU}\left( \sum_{k=1}^K \beta_k^c \cdot A_k(i,j) \right).
\end{equation}
Analogous to $H_{\text{class}}$, we replace the classification confidence $y^c$ with the bounding box regression loss $L_{\text{reg}}$, and obtain activation weights through Huber loss to compute a geometry-aware heatmap $H_{\text{bbox}}$ via gradient backpropagation. Finally, we combine the semantic and geometric information into a mixed heatmap:
\begin{equation}
    H_{\text{mixed}} = \lambda \cdot H_{\text{class}} + (1 - \lambda) \cdot H_{\text{bbox}}.
    \label{eqa:loss}
\end{equation}

\vspace{2mm}\noindent\emph{Step 3.} We introduce a dynamic masking mechanism that converts the heatmap into a binary spatial mask and uses it to modulate positional encodings. Specifically, locations with low heatmap responses ($H_{\text{map}} \leq \tau$) are treated as background-dominant and their positional encodings are suppressed, while locations with high responses ($H_{\text{map}} > \tau$) preserve positional cues for accurate geometric correspondence. The masked positional encodings are then injected into the encoder--decoder pipeline, reducing background-induced positional clutter and improving query retrieval in cluttered scenes.
We formalize the mask filter by thresholding the heatmap:
\begin{equation}
    \text{Mask}(i,j) =
    \begin{cases}
        1, & H_{\text{map}}(i,j) > \tau, \\
        0, & \text{otherwise},
    \end{cases}
    \label{eqa:mask}
\end{equation}
and use it to modulate the standard sinusoidal positional encoding:
\begin{equation}
    \text{PE}(i,j,d) = \text{Mask}(i,j)\odot
    \left[ \sin\left(\frac{i}{\tau_d}\right) + \cos\left(\frac{j}{\tau_d}\right)\right],
\end{equation}
where $\tau_d = 10000^{2d/D}$ and $\odot$ denotes element-wise multiplication.

\vspace{2mm}
The gradient-based heatmap generation (including the second- and third-order derivatives in Eq.~1--2) is used \textit{only during training}. During training, $H_{\text{class}}$ and $H_{\text{bbox}}$ are computed via backpropagation with ground-truth supervision. After convergence, the model retains the learned heatmap-driven embedding behavior in its parameters. At inference time, the detector runs with a standard forward pass \textit{without any gradient computation}, thus incurring no additional runtime overhead beyond conventional transformer inference. This design ensures that the computational cost of higher-order derivatives does not affect deployment efficiency.

\begin{figure}[t] 
    \centering
    \includegraphics[width=0.5\columnwidth]{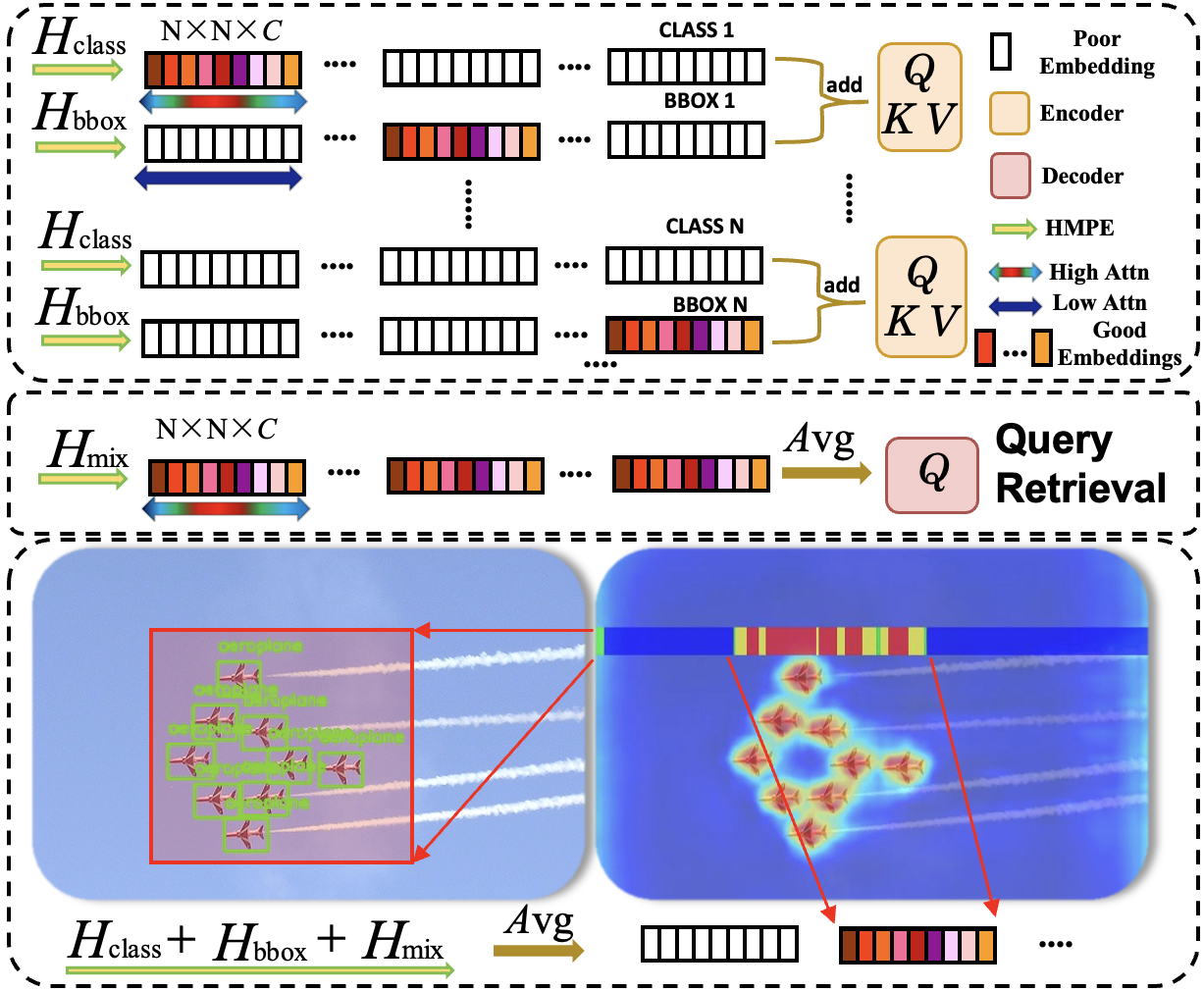}
    \caption{HPE shows where to embed. Top: the encoder fuses heatmap information into embeddings.
    Middle: the decoder retrieves informative embeddings for decoding.
    Bottom: the heatbar visualizes HPE activation (higher = \textcolor{red}{red}, lower = \textcolor{blue}{blue}).}
    \label{fig:vision of HPE}
\end{figure}

HPE uses heatmaps to control positional information: positional encodings are suppressed in background-dominant regions and preserved in foreground-salient regions. As shown in Figure~\ref{fig:vision of HPE}, the heatbar visualization highlights this selective embedding pattern, where hot regions concentrate on detection-relevant areas while cold regions correspond to suppressed background regions. This selective embedding reduces background-induced positional noise and encourages better alignment between positional information and detection semantics.

\begin{figure}[t] 
    \centering
    \includegraphics[width=0.5\columnwidth]{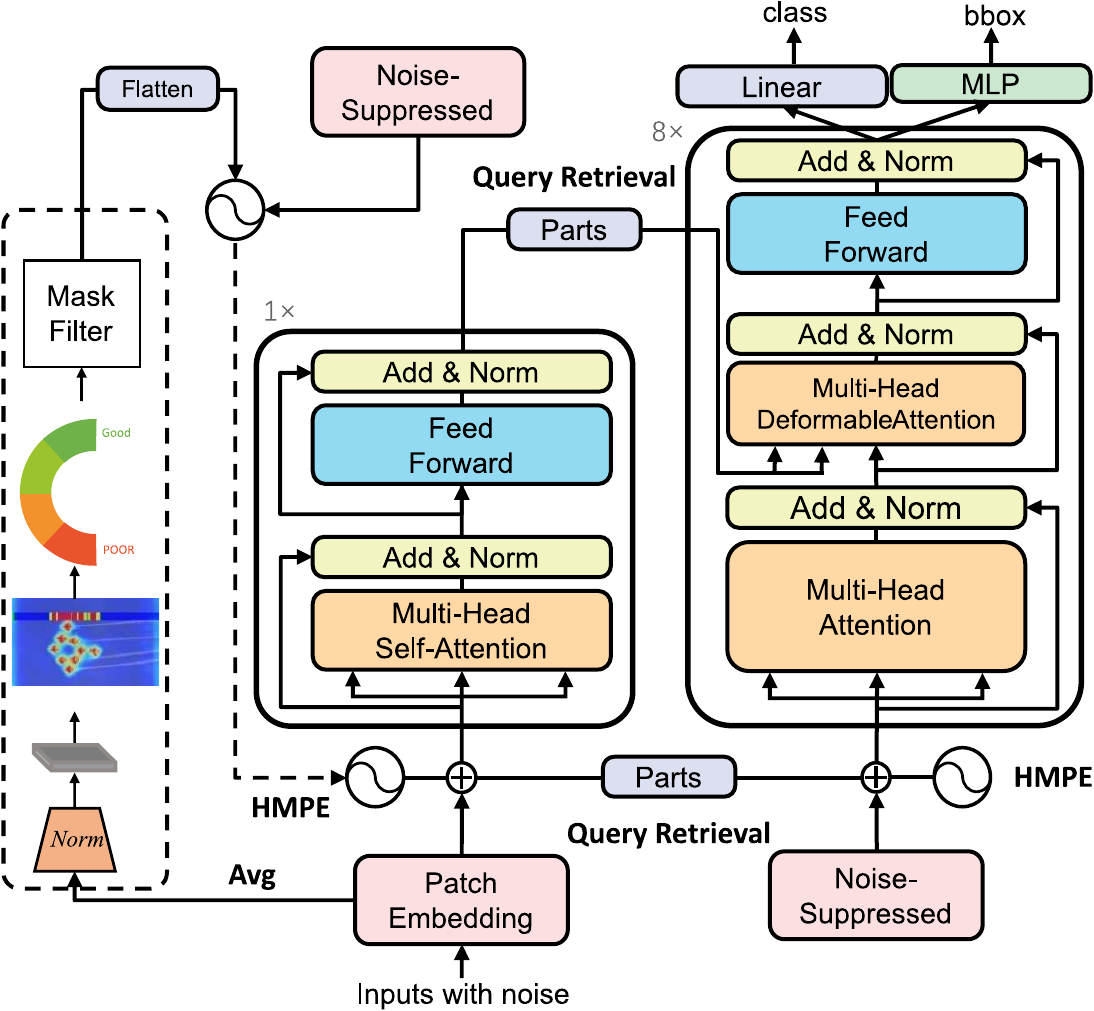}
    \caption{Heatmap-guided query retrieval in an encoder-decoder detector. HPE produces a binary mask from heatmaps and uses it to suppress background-dominant positional embeddings while preserving foreground-salient ones. The masked embeddings are used for noise-suppressed encoding in the encoder and for selecting high-quality queries before decoding.}
    \label{fig:all-crop}
\end{figure}

We adopt hard binarization rather than soft gating (e.g., sigmoid) to enforce a clear foreground/background separation. In aerial imagery, targets are often sparse and the background is large; soft masks may leave residual positional signals in background regions. The binary mask therefore offers a more direct and effective suppression of background positional clutter. Figure~\ref{fig:all-crop} summarizes how the masked positional embeddings are used by the encoder and decoder for query retrieval.

\subsection{Heatmap-Guided Encoding and Retrieval}
\label{subsec:HPE to High-Quality-Query}

This subsection describes how the proposed framework turns the learned positional embeddings into (i) encoder features that are better aligned with detection semantics and (ii) decoder queries that are more discriminative for final prediction. We implement this with two components: MOHFE, which fuses heatmap information into multi-scale encoding and produces the encoder keys/values, and HQ-Retrieval, which converts the mixed-heatmap embedding into a compact set of decoder queries and feeds them into deformable attention for decoding.

\subsubsection{MOHFE: Multi-Scale ObjectBox-Heatmap Fusion Encoder}
\label{subsubsec:MOHFE}

MOHFE is an encoder-side fusion module that turns heatmap-derived positional embeddings into encoder features that are ready for decoder retrieval. Its input consists of two heatmap-conditioned embeddings: a class-driven embedding $E_{\text{class}}$ and a box-driven embedding $E_{\text{bbox}}$. Intuitively, $E_{\text{class}}$ emphasizes category-discriminative regions, while $E_{\text{bbox}}$ emphasizes geometry-related regions; MOHFE combines them so that the encoder feature space carries both semantic and localization cues.

Concretely, we first project these two embeddings and concatenate them into a unified representation $[E_{\text{class}} \| E_{\text{bbox}}]$. We then use separate linear layers to construct the Query/Key/Value tensors for multi-head self-attention in the encoder:
\begin{equation}
    \begin{aligned}
        Q_{\text{enc}} &= W_Q [E_{\text{class}} \| E_{\text{bbox}}], \\
        K_{\text{enc}} &= W_K [E_{\text{class}} \| E_{\text{bbox}}], \\
        V_{\text{enc}} &= W_V [E_{\text{class}} \| E_{\text{bbox}}].
    \end{aligned}
\end{equation}

The encoder attention is performed over these heatmap conditioned features, producing a representation in which foreground-salient locations are more informative for downstream decoding.
When forming the encoder inputs, we use the masked positional encoding defined in Eq.~\eqref{eqa:mask} so that background-dominant positions contribute less positional signal.
As a result, the encoder outputs ($K_{\text{enc}},V_{\text{enc}}$) provide cleaner memory for the decoder-side query retrieval module described next.

\subsubsection{HQ-Retrieval: Heatmap Induced High-Quality Query Retrieval for Decoder}
\label{subsubsec:HQ-Retrieval}

HQ-Retrieval is a decoder-side module that converts the heatmap conditioned embedding from the mixed heatmap $H_{\text{mixed}}$ into a compact set of discriminative decoder queries. Intuitively, $H_{\text{mixed}}$ highlights locations that are simultaneously category-relevant and geometry-consistent; using it as the query source helps avoid decoding from background-dominant embeddings.

Concretely, we first apply a linear projection to obtain the initial query vectors: $Q_{\text{ DeNoise}} = W'_Q E_{\text{mixed}}$. These queries then attend to the encoder memory through deformable attention:
\begin{equation}
    \mathit{DeformAttn}(Q_{\text{DeNoise}}, K_{\text{enc}}, V_{\text{enc}}).
\end{equation}
Because $Q_{\text{DeNoise}}$ is derived from heatmap-guided embeddings that encode both semantics and location, it prioritizes foreground-salient positions and suppresses background-driven responses before decoding. This yields cleaner query-memory interactions, improving both localization and classification while reducing the reliance on repeatedly refining noisy queries with deep decoders.

\begin{figure}[t]
    \centering
    \includegraphics[width=0.9\columnwidth]{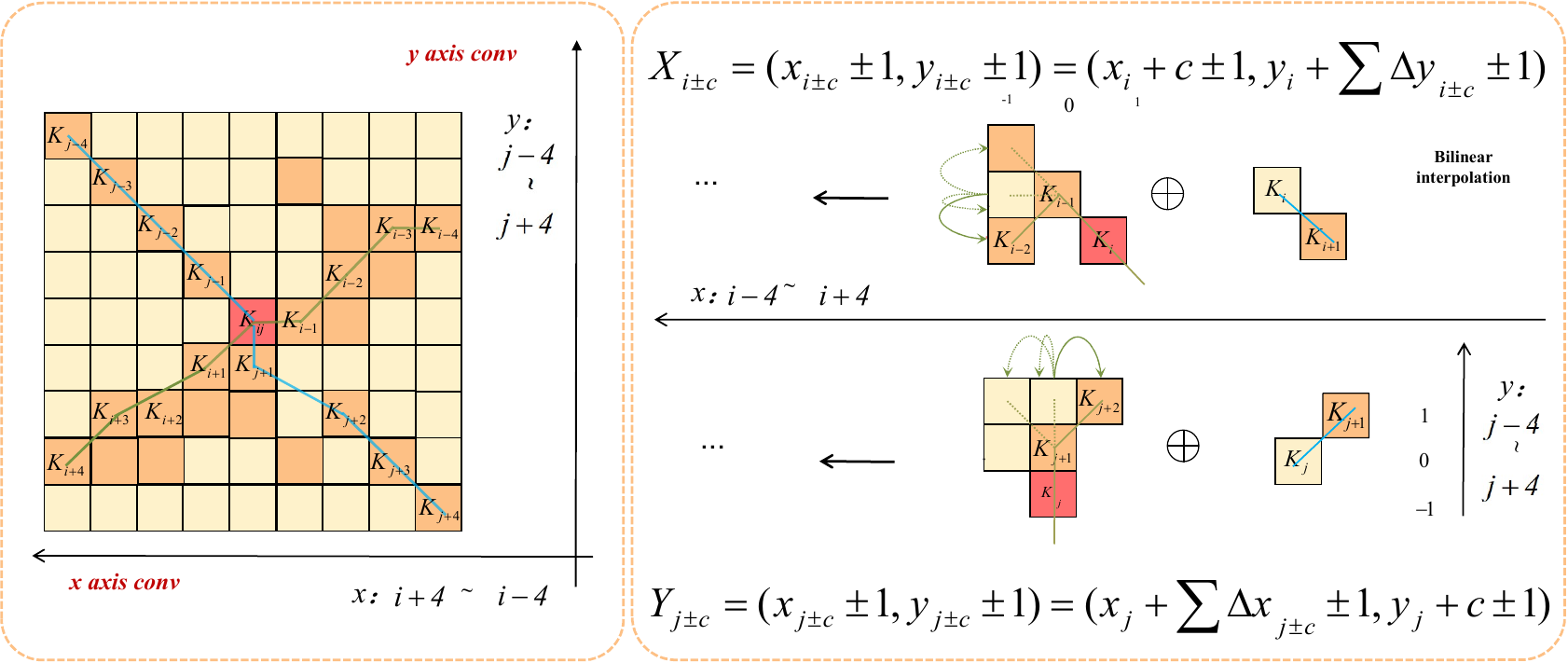}
    \caption{Dual-path axis-wise sampling in LSConv. Left: x-/y-axis convolution samples on a $9\times 9$ grid with linear (\textcolor{blue}{blue}) and snake (\textcolor{green}{green}) paths centered at the red cell. Right: axis-wise offset accumulation and bilinear interpolation used to obtain deformed sampling locations for the two paths.}
    \label{fig:axis-crop}
\end{figure}

\subsection{Linear-Snake Convolution}
\label{subsec:Linear-Snake}

Small-object detection often suffers from extreme feature sparsity: useful evidence may appear as thin, fragmented responses, and a single fixed receptive field can easily miss these weak but structured cues. To strengthen the retrieval-relevant representation before decoding, we introduce Linear-Snake Convolution (LSConv), a lightweight operator that enriches local features with geometry-aware sampling while keeping the sampling behavior stable. Given an input feature map, LSConv outputs an enhanced feature map of the same resolution, which provides a stronger basis for subsequent heatmap-aligned query retrieval.

As shown in Figure~\ref{fig:axis-crop}, LSConv adopts a dual-path design:
(i) a snake branch that follows irregular or curvilinear structures using learnable offsets, and
(ii) a linear branch that enforces straight-line continuity using constrained sampling.
This complementarity lets LSConv capture both ``non-rigid'' local evidence (useful for tiny, cluttered objects) and ``rigid'' linear context (useful for consistent structures), which produces denser and more reliable features for subsequent query retrieval.

LSConv predicts deformation offsets $\Delta$ using a small convolutional predictor. To avoid unstable receptive-field drifting, we constrain the offsets by a continuity regularization so that neighboring sampling locations change smoothly, preventing excessive deformation that could harm alignment. For clarity, consider a $3\times 3$ kernel operating on a $9\times 9$ neighborhood. Along the horizontal direction, LSConv uses a $3\times 1$ strip to aggregate features while keeping the orthogonal direction stable; the deformed sampling coordinates are:
\begin{equation}
    X_{i \pm c} =
    \begin{cases}
        (x_{i+c} + 1,\; y_{i+c} + 1)
        = \left(x_i + c + 1,\; y_i + \sum \Delta y_{i+c} + 1\right), \\
        (x_{i-c} - 1,\; y_{i-c} - 1)
        = \left(x_i - c - 1,\; y_i + \sum \Delta y_{i-c} - 1\right),
    \end{cases}
\end{equation}
where $c \in \{0,1,2,3,4\}$ indexes the distance to the center (Figure~\ref{fig:axis-crop}, right). We sample feature values at these (generally fractional) locations via bilinear interpolation, which keeps LSConv differentiable and stable.

The vertical direction is defined analogously (Figure~\ref{fig:axis-crop}, right):
\begin{equation}
    Y_{j \pm c} =
    \begin{cases}
        (x_{j + c} + 1, y_{j + c} + 1)
        = \left(x_j + \sum \Delta x_{j + c} + 1, \; y_j + c + 1\right), \\
        (x_{j - c} - 1, y_{j - c} - 1)
        = \left(x_j + \sum \Delta x_{j - c} - 1, \; y_j + c - 1\right).
    \end{cases}
\end{equation}

\begin{figure}[t]
    \centering
    \includegraphics[width=0.6\columnwidth]{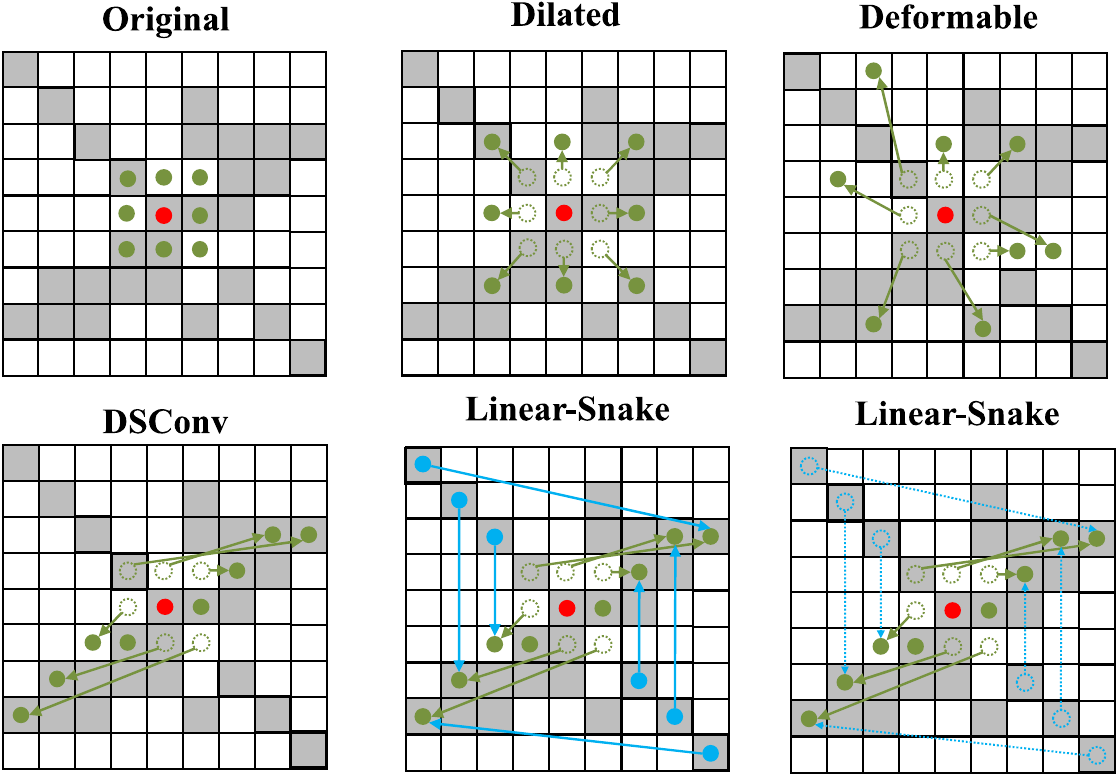}
    \caption{Sampling patterns comparison on a $9\times 9$ grid. We contrast standard, dilated, deformable~\cite{zhu2019deformable}, and DSC~\cite{qi2023dynamic} convolutions with two LSConv variants. LSConv combines a constrained linear path and a deformable snake-like path to cover both straight and curved structures, while other operators follow their fixed or unconstrained sampling layouts.}
    \label{fig:conv-crop}
\end{figure}

\begin{table*}[htbp]
    \centering
    \caption{Comparison performance on PASCAL VOC and NWPU VHR-10. Compared with RT-DETR, our method achieves consistently higher accuracy on both datasets while markedly reducing computational and model complexity. The best and second-best results are highlighted in \textcolor{red}{red} and \textcolor{blue}{blue}.}
    \label{tab:comparison}
    \renewcommand{\arraystretch}{1.2}
    \setlength{\tabcolsep}{3pt}
    \resizebox{0.85\textwidth}{!}{%
    \begin{tabular}{l l c c c c c c c c c}
        \hline \hline
        \multirow{3}{*}{ } & \multirow{3}{*}{Method} & \multirow{3}{*}{Date} & \multirow{3}{*}{Backbone} & \multicolumn{4}{c}{Datasets} & \multirow{3}{*}{Epoch} & \multirow{3}{*}{GFLOPs} & \multirow{3}{*}{\begin{tabular}{c} Params \\ (M) \end{tabular}} \\
        \cline{5-8}
        & & & & \multicolumn{2}{c}{PASCAL VOC} & \multicolumn{2}{c}{NWPU} & & & \\
        \cline{5-8}
        & & & & mAP@0.5 & mAP & mAP@0.5 & mAP & & & \\
        \hline
        \multirow{14}{*}{CNN} & SSD & 2016ECCV & RestNet18 & 51.01 & 22.98 & 40.34 & 16.01 & 75 & 353 & 26.29 \\
        & FCOS & 2019ICCV & ResNet18 & 67.41 & 43.64 & 84.35 & 55.68 & 75 & 32 & 31.8 \\
        & RetinaNet & 2018CVPR & ResNet18 & 58.21 & 44.50 & 87.78 & 57.89 &75 & 39 & 36.1 \\
        & Faster R-CNN & 2017ICCV & VGG16 & 65.97 & 38.63 & 89.15 & 61.23 &75 & 63 & 41.1 \\
        & CenterNet & 2019ICCV & ResNet34 & 57.64 & 30.90 & 81.28 & 48.76 & 75 & 130 & 41.7 \\
        & MobilenetV3 & 2019ICCV & MBNV3 & 52.40 & 30.12 & 62.97 & 40.56 & 75 & 14 & 8.86 \\
        & YOLOv3 & 2018CVPR & ResNet18 & 64.08 & 43.76 & 84.50 & 52.34 & 75 & 112 & 43.1 \\
        & YOLOv4 & 2020CVPR & Darknet-53 & 62.70 & 49.90 & 87.80 & 60.89 & 75 & 101 & 44.9 \\
        & YOLOv5 & 2020ArXiv & CSPDarknet53 & 67.10 & 49.73 & 89.11 & 65.43 & 75 & 109 & 46 \\
        & YOLOv6 & 2022ArXiv & CSPDarknet53 & 67.16 & 54.28 & 89.78 & 52.10 &75 & 150 & 59 \\
        & YOLOv7 & 2023CVPR & CSPRep53 & 67.80 & 44.30 & 91.03 & 63.12 &75 & 104 & 36 \\
        \multirow{10}{*}{Transformer} & YOLOv8 & 2023Arxiv & CSPDarknet18 & 53.01 & 36.66 & 90.43 & 60.80 & 75 & 32 & 16.1 \\
        & YOLOv11 & 2025ArXiv & CSPRep53 & \textcolor{red}{71.70} & 47.3 & \textcolor{blue}{92.68} &\textcolor{red}{ 68.89} & 75 & 105 & 38.8 \\
        \hline
        
        & DETR & 2020ECCV & ResNet18 & 62.40 & 42.00 & 81.28 & 35.9 & 125 & 101 & 36.74 \\
        & Deformable DETR & 2021ICLR & ResNet50 & 60.56 & \textcolor{red}{51.92} & 79.30 & 40.50 &125 & 179 & 39.83 \\
        & Conditional-DETR & 2021ICCV & ResNet50 & 68.35 & 48.72 & 91.45 & 64.38 & 100 & 98 & 240.2 \\
        & DINO & 2023ICLR & ResNet50 & 67.92 & 45.16 & 90.87 & 62.95 & 12 & 112 & 247.5 \\
        & PR-Deformable DETR* & 2025GRSL & ResNet50 & / & / & 88.30 & 43.20 & 125 & 151 & 46.07 \\
        & RT-DETR & 2024CVPR & HGNetv1 & 69.41 & 50.59 & 92.60 & 60.26 & 100 & 136 & 163 \\
        \rowcolor{LightGreen} 
        & Ours & 2026 & HPE-Bone & \textcolor{blue}{70.60} & \textcolor{blue}{51.59} & \textcolor{red}{94.51} & \textcolor{blue}{67.20} & 100 & 57 & 66.3 \\
        \hline \hline
    \end{tabular}%
    }
\end{table*}

The snake branch dominates when evidence is fragmented or curved (e.g., small objects under clutter), because flexible offsets can ``trace'' the local response and recover missing context. The linear branch dominates when the scene contains stable straight structures, because constrained sampling preserves consistent geometry and avoids over-deformation. Figure~\ref{fig:conv-crop} further contrasts LSConv with standard, dilated, deformable, and DSC operators: LSConv explicitly couples a constrained linear sampler with a deformable snake sampler, rather than relying on a single sampling rule. In practice, we insert LSConv into the backbone to enhance feature maps before they are passed to the subsequent HPE-driven encoding and query retrieval modules. Overall, LSConv provides a simple but effective way to densify sparse information and improve geometric consistency, which makes the features more suitable for heatmap-aligned query retrieval in small-object scenarios.

\section{Experimental Results and Analysis}

\subsection{Datasets and Experimental Protocol}
We evaluate our method on five public benchmark datasets covering generic and aerial object detection: NWPU VHR-10~\cite{cheng2016learning} (aerial imagery with small and densely distributed objects), PASCAL VOC~\cite{everingham2010pascal} (generic object detection), and three large-scale aerial datasets (DOTA, DIOR, and VisDrone)~\cite{yu2025visualizing,yu2025iidm,yu2026dinov3,yu2026spatiotemporal} to assess scalability across data types and scene complexity.

For all datasets, we follow the official training/testing splits and strictly adopt the evaluation protocol in~\cite{zhao2024detrs}. Unless otherwise stated, all hyperparameters are kept identical across datasets. In particular, we set $\lambda=0.5$ in Eq.~\eqref{eqa:loss} to balance semantic and geometric priors, and use $\tau=0.5$ in Eq.~\eqref{eqa:mask} to binarize the heatmap for subsequent query retrieval. Both $\lambda$ and $\tau$ are selected on the validation set and then fixed for all benchmarks. We report mAP@50 and mAP as detection accuracy metrics, and use GFLOPs and Params (M) to measure computational cost and model size, respectively.

\begin{table*}[t]
\centering
    \caption{Ablation study of our design components on NWPU VHR-10 and PASCAL VOC. Each component contributes incremental improvements, and the full configuration consistently achieves the best performance on both datasets.}
    \label{tab:ablation}
    \adjustbox{max width=0.85\textwidth}{
        \begin{tabular}{c c c c c|c c|c c}
            \toprule
            \toprule
            \multirow{2}{*}{Dataset} 
            & \multirow{2}{*}{DSconv} 
            & \multirow{2}{*}{Linear-Snake} 
            & \multicolumn{2}{c|}{Heatmap-guided Positional Embedding} 
            & \multirow{2}{*}{Precision(\%)} 
            & \multirow{2}{*}{Recall(\%)} 
            & \multirow{2}{*}{mAP(\%)} 
            & \multirow{2}{*}{mAP@95(\%)} \\
            \cmidrule(lr){4-5}
            & & & MOHFE & HQ-Retrieval & & & & \\
            \midrule

            \multirow{5}{*}{NWPU VHR-10}
            & - & - & - & - 
            & 88.9 & 89.9 & 92.6 & 60.2 \\
            & $\checkmark$ & - & - & - 
            & 89.0$_{\uparrow 0.10}$ & 90.10$_{\uparrow 0.20}$ & 92.71$_{\uparrow 0.11}$ & 62.32$_{\uparrow 2.12}$ \\
            & $\checkmark$ & $\checkmark$ & - & - 
            & 90.74$_{\uparrow 1.84}$ & 90.62$_{\uparrow 0.72}$ & 93.91$_{\uparrow 1.31}$ & 63.69$_{\uparrow 3.49}$ \\
            & $\checkmark$ & $\checkmark$ & $\checkmark$ & - 
            & 90.75$_{\uparrow 1.85}$ & 90.51$_{\uparrow 0.61}$ & 94.18$_{\uparrow 1.58}$ & 63.94$_{\uparrow 3.74}$ \\
            \rowcolor{LightGreen}
            & $\checkmark$ & $\checkmark$ & $\checkmark$ & $\checkmark$ 
            & 93.17$_{\uparrow 4.97}$ & 90.40$_{\uparrow 0.50}$ & 94.51$_{\uparrow 1.91}$ & 67.20$_{\uparrow 6.90}$ \\
            \midrule

            \multirow{5}{*}{PASCAL VOC}
            & - & - & - & - 
            & 76.7 & 62.9 & 69.4 & 50.5 \\
            & $\checkmark$ & - & - & - 
            & 77.20$_{\uparrow 0.50}$ & 62.93$_{\uparrow 0.03}$ & 70.29$_{\uparrow 0.89}$ & 51.17$_{\uparrow 0.67}$ \\
            & $\checkmark$ & $\checkmark$ & - & - 
            & 78.41$_{\uparrow 1.71}$ & 63.10$_{\uparrow 0.20}$ & 70.19$_{\uparrow 0.79}$ & 51.41$_{\uparrow 0.91}$ \\
            & $\checkmark$ & $\checkmark$ & $\checkmark$ & - 
            & 78.52$_{\uparrow 1.82}$ & 63.55$_{\uparrow 0.65}$ & 70.47$_{\uparrow 1.07}$ & 51.33$_{\uparrow 0.83}$ \\
            \rowcolor{LightGreen}
            & $\checkmark$ & $\checkmark$ & $\checkmark$ & $\checkmark$ 
            & 78.80$_{\uparrow 2.10}$ & 63.72$_{\uparrow 0.82}$ & 70.60$_{\uparrow 1.20}$ & 51.59$_{\uparrow 1.09}$ \\

            \bottomrule
            \bottomrule
        \end{tabular}
    }
\end{table*}

\begin{figure*}[t] 
    \centering
    \includegraphics[width=0.76\textwidth]{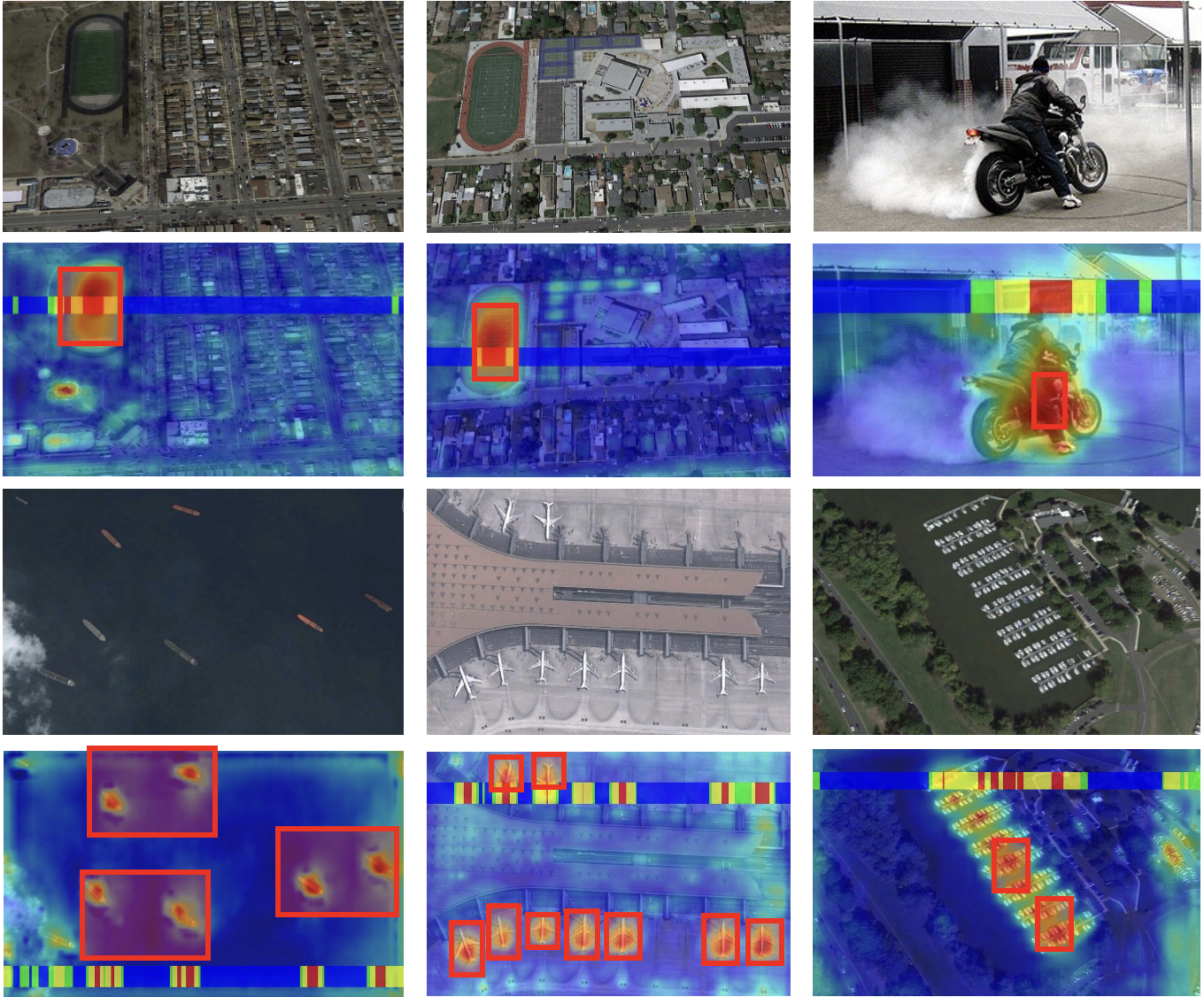}
    \caption{Visualization of HPE heatbars under cluttered versus clean scenes. The heatmap guidance provides more object-aware spatial information, leading to tighter and more stable localization under background distractions and scale variations.}
    \label{fig:vis2}
\end{figure*}

\subsection{Result Analysis}
\subsubsection{Quantitative Results}
We compare with representative CNN-based detectors, including FCOS~\cite{tian2019fcos}, RetinaNet~\cite{lin2017focal}, Faster R-CNN~\cite{ren2015faster}, CenterNet~\cite{duan2019centernet}, MobileNet-V3~\cite{howard2019searching}, and the YOLO series~\cite{redmon2018yolov3,bochkovskiy2020yolov4,benjumea2021yolo,li2022yolov6,wang2023yolov7,khanam2024yolov11}. We also include state-of-the-art Transformer detectors (DETR, Deformable DETR, PR-Deformable DETR*, and RT-DETR) to benchmark against encoder-decoder paradigms most closely related to our design. For fairness, CNN baselines are trained with standardized hyperparameters~\cite{wang2021footprint,yu2026ads,yu2026cast,zeng2025hmpe,zeng2026deepinterestgr}, while Transformer-based baselines are configured to match our experimental setting and their commonly used training schedules.

Table~\ref{tab:comparison} summarizes the results on PASCAL VOC and NWPU VHR-10. CNN detectors typically reach convergence within shorter schedules (e.g., 75 epochs), whereas Transformer detectors often require longer training (100--125 epochs) to stabilize optimization and reach competitive performance. This behavior is particularly pronounced on NWPU VHR-10, where small objects and limited training data make query learning more challenging for Transformer-based detectors (e.g., DETR yields 35.9\% mAP).

Our method achieves 94.51\% mAP@0.5 and 67.20\% mAP on NWPU VHR-10. Relative to the strongest real-time Transformer baseline RT-DETR, we improve both accuracy and efficiency: +1.91 mAP@0.5 and +6.94 mAP, while reducing computation from 136 to 57 GFLOPs and parameters from 163M to 66.3M (59.4\% fewer parameters). These results indicate that our heatmap-guided positional embedding effectively stabilizes high-quality query retrieval in encoder-decoder detection and yields a better accuracy-efficiency trade-off under small-object and sparse-data regimes.

\subsubsection{Qualitative Visualization}

Figure~\ref{fig:vis2} provides example visualizations of the learned HPE patterns for denoised query retrieval on PASCAL VOC and NWPU VHR-10. The figure is arranged in four rows: the odd rows show the input images, and the even rows overlay the corresponding HPE heatbars on the original resolution. These examples consistently exhibit a ``hot-center, cold-ends'' profile, which confirms the core design of HPE: injecting positional information that emphasize object-relevant regions while suppressing background responses, thereby reducing noisy query proposals at initialization. Moreover, the visualizations suggest that HPE is not merely a saliency-like highlight; rather, it induces a structured spatial prior that favors compact, unimodal activations and sharper spatial contrast. This property is particularly aligned with high-IoU evaluation, as it promotes more stable query anchoring and tighter localization under cluttered scenes or scale variations. Minor spatial offsets may appear due to padding and resizing in the downsampling/upsampling path; however, they do not change the qualitative activation reminder that HPE concentrates on object-centric regions.

\subsection{Ablation Studies and Additional Evaluation}

\subsubsection{Component-wise ablation}
Table~\ref{tab:ablation} reports a component-wise ablation of our framework on NWPU VHR-10 and PASCAL VOC. Overall, enabling modules progressively improves both datasets, and the full configuration consistently achieves the best performance, suggesting that the proposed components contribute complementary gains rather than redundant modifications.

\noindent\textbf{NWPU VHR-10.}
The baseline model achieves 92.6\% mAP@0.5 and 60.2\% mAP. Adding DSconv yields modest but consistent gains (+0.11 mAP@0.5 and +2.12 mAP), indicating improved feature stability for small and densely distributed objects. Introducing Linear-Snake further enhances geometric modeling, improving performance to 93.91\% mAP@0.5 and 63.69\% mAP, which suggests better sensitivity to object boundaries and shape information. With MOHFE enabled, mAP@0.5 increases to 94.18\% and mAP to 63.94\%, validating that heatmap-guided positional encoding provides a cleaner spatial prior. Finally, integrating HQ-Retrieval leads to the most significant improvement, boosting mAP@0.5 to 94.51\% and mAP to 67.20\% (+6.9 over the baseline). Notably, the gain is substantially larger on mAP than on mAP@0.5, confirming that these components primarily improve high-IoU localization quality by refining decoder queries toward geometrically consistent regions.

\noindent\textbf{PASCAL VOC.}
A similar trend is observed on PASCAL VOC. The full model improves mAP@0.5 from 69.4\% to 70.6\% and mAP from 50.5\% to 51.6\%. Although the absolute gains are smaller than those on NWPU VHR-10, the consistent improvements across both metrics indicate that these components generalize beyond aerial imagery and remain effective under generic object detection settings.

\subsubsection{Ablation on DETR decoder depth}


\begin{table}[t]
\setlength{\aboverulesep}{0.5ex}
\setlength{\belowrulesep}{0.5ex}
\renewcommand{\arraystretch}{1.2}

\caption{Decoder multi-head ablation under different training regimes. Det$^k$ denotes using $k$ decoder layers for detection, while Det$^{3\text{--}7}$ aggregates outputs from layers 3 to 7, where performance saturates. Results show that increasing decoder depth consistently improves AP under limited training budgets, but exhibits diminishing returns beyond two layers under sufficient training.}
\label{tab:params}
\centering

\resizebox{0.5\linewidth}{!}{%
\begin{tabular}{c|c c c c|c c c c}
    \toprule
    \toprule
    \multirow{2}{*}{Epoch} & \multicolumn{4}{c}{AP(\%)} & \multicolumn{4}{c}{GFLOPs} \\
    \cline{2-5} \cline{6-9}
    & Det$^0$ & Det$^1$ & Det$^2$ & Det$^{3\text{--}7}$ & 0 & 1 & 2 & 3\text{--}7 \\
    \midrule
    100 & 66.7 & 69.9 & \textcolor{red}{70.5} & \textcolor{red}{70.5}
        & \multirow{4}{*}{53.7} & \multirow{4}{*}{55.4} & \multirow{4}{*}{56.2} & \multirow{4}{*}{57.0} \\
    50  & 62.4 & 65.6 & 63.1 & \textcolor{red}{66.8} &  &  &  &  \\
    25  & 40.7 & 41.3 & 43.2 & \textcolor{red}{47.9} &  &  &  &  \\
    Early-stop & 38.9 & 42.9 & 43.8 & \textcolor{red}{43.9} &  &  &  &  \\
    \bottomrule
    \bottomrule
\end{tabular}%
}
\end{table}

Table~\ref{tab:params} analyzes the effect of decoder depth in a DETR-style architecture under different training regimes. All configurations employ the same HPE-driven query augmentation, and only the number of decoder layers used for detection varies. When sufficient training is available (100 epochs), Det$^{2}$ already reaches peak performance (70.5\% AP), and deeper configurations (Det$^{3\text{--}7}$) do not yield further improvement, indicating clear performance saturation. Under limited training budgets, however, deeper decoders are beneficial: at 25 epochs and in the early-stop regime, Det$^{3\text{--}7}$ consistently outperforms shallower variants. Here, ``early-stop'' denotes a setting ($<10$ epochs) that terminates once the training dynamics stabilize.

These results suggest that HPE improves query quality, thereby reducing the reliance on deep decoder stacks. In practice, two decoder layers are sufficient under full supervision, while three layers provide a favorable trade-off in data-scarce regimes. The corresponding GFLOPs increase across decoder depths is modest (53.7$\rightarrow$57.0), reflecting the encoder-dominated computation pattern of DETR-like architectures; as a result, depth reduction tends to yield more noticeable savings in parameters and latency than in raw FLOPs. Finally, the encoder--decoder parameter split (3.01\,MB vs.\ 16.8\,MB) indicates a design that allocates more capacity to query refinement than to feature extraction, which aligns with our objective of precise localization via high-quality retrieval.

\subsubsection{Additional Evaluation on Benchmarks}
\label{subsec:large-scale}

\begin{table}[t]
    \centering
    \caption{Generalization to large-scale aerial benchmarks. We report AP at IoU thresholds 0.50 and 0.75 (AP$_{50}$ / AP$_{75}$). ``Ours'' consistently improves over the DETR across all datasets.}
    \label{tab:aerial}
    \renewcommand{\arraystretch}{1.3}
    \resizebox{0.45\columnwidth}{!}{
        \begin{tabular}{l c c c c c c}
            \toprule
            \toprule
            \multirow{2}{*}{Method} 
            & \multicolumn{2}{c}{DOTA} 
            & \multicolumn{2}{c}{DIOR} 
            & \multicolumn{2}{c}{VisDrone} \\
            \cmidrule(lr){2-3} \cmidrule(lr){4-5} \cmidrule(lr){6-7}
            & AP$_{50}$ & AP$_{75}$ 
            & AP$_{50}$ & AP$_{75}$ 
            & AP$_{50}$ & AP$_{75}$ \\
            \midrule
            DETR & 33.7 & 15.4 & 56.2 & 25.8 & 42.1 & 19.7 \\
            \rowcolor{LightGreen} 
            Ours & 40.1$_{\uparrow 6.4}$ & 19.3$_{\uparrow 3.9}$ 
                 & 63.8$_{\uparrow 7.6}$ & 31.2$_{\uparrow 5.4}$ 
                 & 49.6$_{\uparrow 7.5}$ & 24.1$_{\uparrow 4.4}$ \\
            \bottomrule
            \bottomrule
        \end{tabular}
    }
\end{table}

We further examine the generalizability of our method on three large-scale aerial detection benchmarks (DOTA, DIOR, and VisDrone) by comparing against a DETR baseline. Table~\ref{tab:aerial} shows that our method consistently outperforms DETR on all three benchmarks at both AP$_{50}$ and AP$_{75}$. The improvements are +6.4/+3.9 on DOTA, +7.6/+5.4 on DIOR, and +7.5/+4.4 on VisDrone (AP$_{50}$/AP$_{75}$), indicating that the HPE-based query retrieval mechanism transfers reliably to large-scale aerial datasets with different object densities and scale distributions.

\section{Conclusion}

We study a simple but under-explored question for query-based small-object detection: \emph{where to embed} positional information. In cluttered aerial scenes, uniformly injecting positional embeddings lets background-dominant regions carry distracting signals, which degrades retrieval and shifts the burden to deeper decoders for repeated refinement. By making positional embedding selective and preserving it in foreground-salient regions while suppressing it elsewhere, our method yields cleaner and more stable decoder inputs, improves high-IoU localization, and reduces the reliance on deep decoder stacks. This leads to a practical accuracy and efficiency gain, where the decoder can be substantially lightened without sacrificing performance. Overall, the key takeaway is that \emph{where} positional embedding is applied matters as much as having it, and viewing embedding placement as a noise-aware allocation problem provides a simple route to both stronger accuracy and a more efficient detector.
Future work includes studying prediction-driven alternatives to gradient-based saliency for positional allocation and examining how selective positional embedding generalizes to other query-based detection architectures and datasets.

\bibliographystyle{unsrt}  
\bibliography{refs}

\end{document}